\documentclass[]{article} 

\usepackage{aaai21}  
\usepackage{times}  
\usepackage{helvet} 
\usepackage{courier}  
\usepackage[hyphens]{url}  
\usepackage{graphicx} 
\usepackage{natbib}  
\usepackage{caption} 
\usepackage{multirow}
\usepackage{booktabs}
\usepackage{makecell}
\usepackage{subcaption}
\usepackage{amsmath}
\usepackage{booktabs}
\usepackage{arydshln}
\usepackage{color}
\usepackage{enumerate}
\usepackage{tabularx}
\usepackage[switch]{lineno}
\usepackage{xspace}

\usepackage[normalem]{ulem}

\usepackage[dvipsnames]{xcolor}

\usepackage{bbm}

\definecolor{applegreen}{rgb}{0.55, 0.71, 0.0}

\nocopyright
\title{UniDS : A Unified Dialogue System for Chit-Chat and Task-oriented Dialogues}
\author {
    Xinyan Zhao\textsuperscript{\rm 1},
    Bin He\textsuperscript{\rm 2},
    Yasheng Wang\textsuperscript{\rm 2},
    Yitong Li\textsuperscript{\rm 2},
    Fei Mi\textsuperscript{\rm 2},
    Yajiao Liu\textsuperscript{\rm 2},\\
    Xin Jiang\textsuperscript{\rm 2},
    Qun Liu\textsuperscript{\rm 2},
    Huanhuan Chen\textsuperscript{\rm 1}
    \\
}
\affiliations {
    \textsuperscript{\rm 1} University of Science and Technology of China\\
    \textsuperscript{\rm 2} Huawei Noah’s Ark Lab\\
    sa516458@mail.ustc.edu.cn, \{hebin.nlp,\ wangyasheng,\ liyitong3,\ mifei2,\ yajiao.liu,\ Jiang.Xin,\ qun.liu\}@huawei.com, hchen@ustc.edu.cn
}

\begin{document}
\maketitle

\begin{abstract}
With the advances in deep learning, tremendous progress has been made with chit-chat dialogue systems and task-oriented dialogue systems. However, these two systems are often tackled separately in current methods. To achieve more natural interaction with humans, a dialogue agent needs to be capable of both chatting and accomplishing tasks. To this end, we propose a unified dialogue system (UniDS) with the two aforementioned skills.
In particular, we design a unified dialogue data schema, compatible for both chit-chat and task-oriented dialogues, and we train UniDS with mixed dialogue data from a pretrained chit-chat dialogue model. Without adding extra parameters to SOTA baselines, UniDS can alternatively handle chit-chat and task-oriented dialogues in a unified framework. Experimental results demonstrate that the proposed UniDS works comparably well as the pure chit-chat system, and it outperforms state-of-the-art task-oriented dialogue systems. More importantly, UniDS achieves better robustness as it is able to smoothly switch between two types of dialogues.
These results demonstrate the feasibility and potential of building an \textit{one-for-all} dialogue system.
\end{abstract}

\section{Introduction}
\begin{figure}[t]
\centering
\framebox{\includegraphics[width=0.43\textwidth]{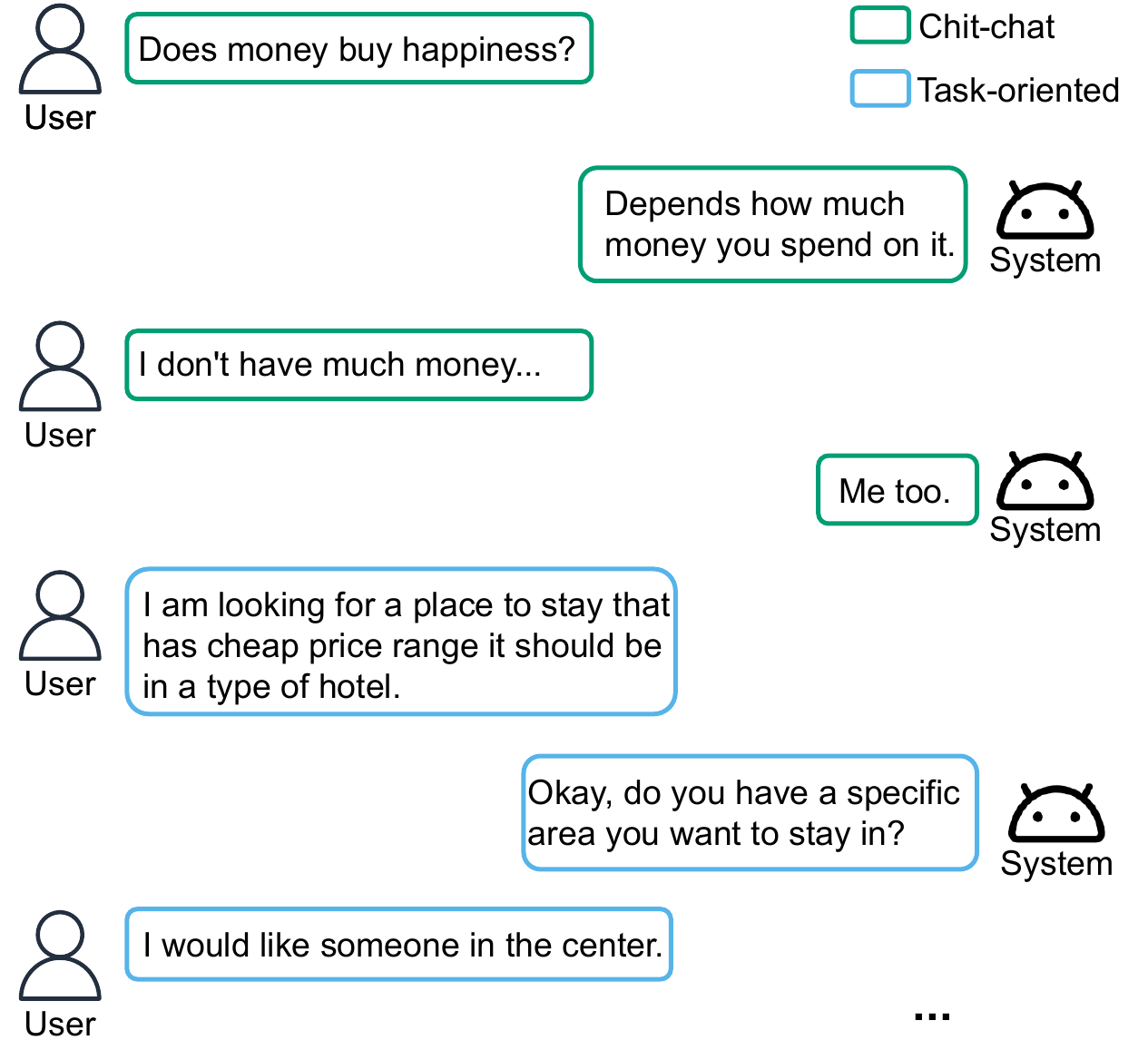}} 
\caption{Illustration of users being interested to chit-chat with the dialogue system before booking a hotel.}
\label{dowhat}
\end{figure}

Dialogue system is an important tool to achieve intelligent user interaction, and it is actively studied by NLP and other communities. Current research of dialogue systems focus on task-oriented dialogue (TOD) systems \cite{Hosseini-AslMWY20,abs-2005-05298,YangLQ21}, achieving functional goals, and chit-chat dialogue systems aiming at entertainment \cite{ZhouHZZL18, ZhangSGCBGGLD20, ZhaoXZYC20, RollerDGJWLXOSB21}.
Different methods are devised for these two types of dialogue systems separately. However, a more suitable way for human would be to have one dialogue agent that is able to handle both chit-chat and TOD in one conversation.
As illustrated in Figure~\ref{dowhat}, users may have communication-oriented needs (e.g. chatting about money and happiness) and task-oriented needs (e.g. hotel reservation) when interacting with a dialogue agent.
Furthermore, inputs of dialogue systems are often interfered by background noise, such as voice from other people or devices, collected by the preceding automatic speech recognition (ASR) module.
Therefore, considering the chit-chat ability may also improve the robustness of a task-oriented dialog system \cite{ZhaoLLE17}.

Creating a unified model for different tasks without performance degradation is challenging \cite{KaiserGSVPJU17}. Some works attempt to model different dialogue skills via different experts or adapters \cite{abs-2001-01871, LinMBF21}. However, these methods increase the number of parameters and need to explicitly select dialogue skills. Also, these works lack the exploration of the ability to switch between different types of dialogues.

\begin{figure*}[!t]
\centering
\includegraphics[width=1\textwidth]{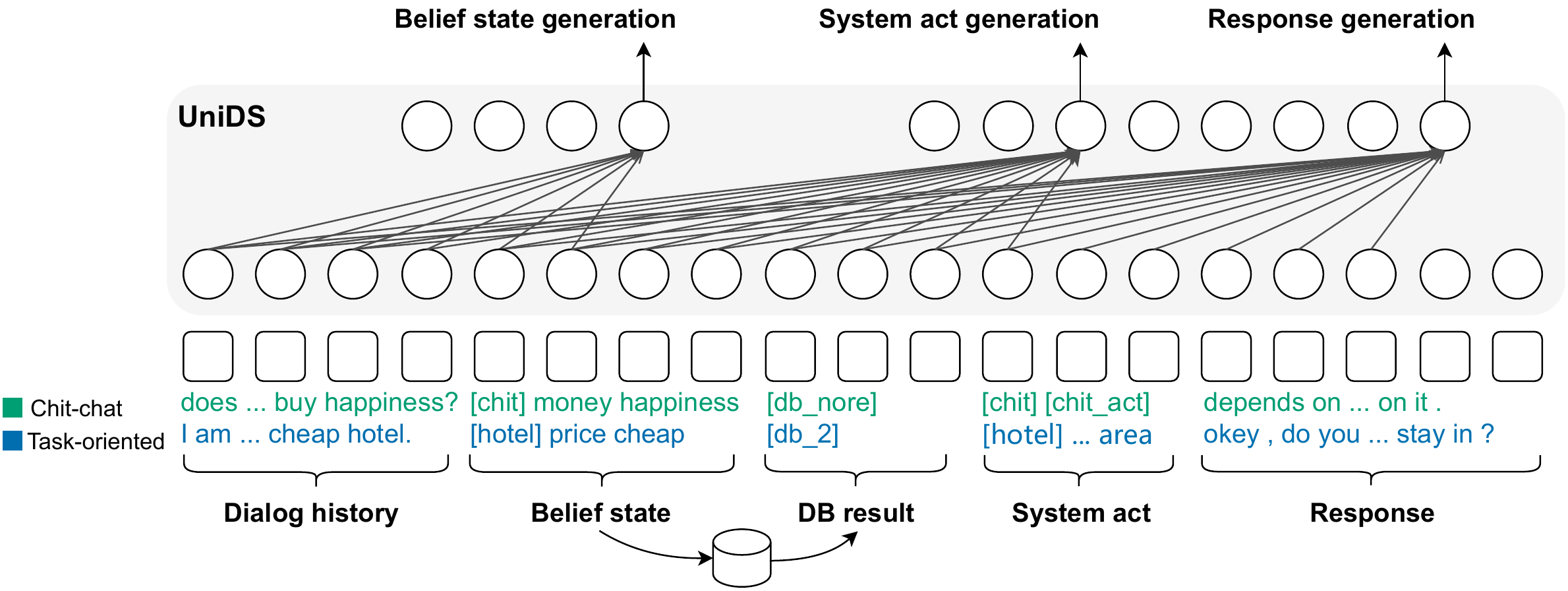} 
\caption{Proposed architecture of Unified dialogue system (UniDS).}
\label{fig:overview}
\end{figure*}

Motivated by the recent success of applying pre-trained language models for task-oriented dialogue systems \cite{Hosseini-AslMWY20,abs-2005-05298,YangLQ21} and chit-chat dialogue systems \cite{ZhangSGCBGGLD20, abs-2001-09977, RollerDGJWLXOSB21, abs-2006-16779},
we propose a pre-training based dialogue system (UniDS) to handle chit-chat and TOD in a unified framework. Specifically, to unify chit-chat and task-oriented dialogues, we device (1) belief state (2) representations of database result, and (3) system act for chit-chat dialogues as in task-oriented dialogues. With this unified \textbf{data schema}, we mix two types of dialogues and train UniDS on the basis of the state-of-the-art chit-chat dialogue system (DialoGPT \cite{ZhangSGCBGGLD20}). 
Moreover, we find that the ``entity recommendation'' act is important for task completion, but it is not given enough credits when training UniDS with mixed dialogues. To address this problem, we propose to utilize a weighted cross-entropy loss to give more attention to the entity recommendation act. 

We evaluate UniDS using a public task-oriented dialogue dataset MultiWOZ and an 8k chit-chat dataset extracted from Reddit through both automatic and human evaluations.
UniDS achieves comparable performance compared to the state-of-the-art chit-chat dialogue system (DialoGPT), and it outperforms the state-of-the-art TOD system (UBAR; \citet{YangLQ21}).
In addition, we also empirically show that UniDS is more robust to noise in task-oriented dialogues, and UniDS shows a desirable ability to switch between the two types of dialogues.

The contributions of this work are summarised as follows:
\begin{itemize}
  \item 
  To the best of our knowledge, this is the first work presenting a unified dialogue system to jointly handle chit-chat and task-oriented dialogues in an end-to-end way. 
  \item
  We design a unified dialogue \emph{data schema} for TOD and chit-chat, allowing the training and inference of dialogue systems to be performed in a unified manner.
  \item
  Extensive empirical results show that UniDS performs comparably to state-of-the-art chit-chat dialogue systems and outperforms state-of-the-art TOD systems. Moreover, UniDS achieves better robustness to dialog noise and better switchability between two types of dialogues. 

\end{itemize}

\section{Related Work}
With the development of large-scale language models, chit-chat dialogue systems achieve remarkable success. Based on GPT-2 \cite{radford2019language}, DialoGPT \cite{ZhangSGCBGGLD20} is further trained on large-scale dialogues extracted from Reddit. DialoGPT could generate more relevant, contentful, and fluent responses than previous methods. Afterwards, larger pre-train LM based chit-chat dialogue systems \cite{abs-2001-09977,abs-2006-16779,RollerDGJWLXOSB21} are proposed and achieve even better performance. In the area of task-oriented dialogue systems, recent research \cite{Hosseini-AslMWY20,abs-2005-05298, YangLQ21} concatenated elements in a dialogue into one sequence and utilized pre-train LM to generate the belief state, system act, and response in an end-to-end way and achieved promising results.


There are several works related to the unified dialogue system. \citet{ZhaoLLE17} insert one turn chit-chat dialogue into task-oriented dialogues to train a model with better out-of-domain recovery ability.
Attention over Parameters (AoP) \cite{abs-2001-01871} utilizes different decoders for different dialogue skills (e.g., hotel booking, restaurant booking, chit). However, the performance of AoP can be improved and it largely increases parameters comparing with models that handle a single type of dialogues. ACCENTOR \cite{SunMCRSLWLCC21} adds chit-chat utterance at the beginning or end of task-oriented responses to make the conversation more engaging, but ACCENTOR is unable to have a chit-chat with users. Unlike the above works, UniDS does not add extra parameters to existing dialogue models, and UniDS could alternatively handle chit-chat and task-oriented dialogues in a seamless way.

\section{Unified Dialogue System}

\begin{table*}[!t]
\centering
\begin{tabular}{llll}
    \toprule
    & \bf{Unified dialogue data schema} &\bf{Chit-chat example} &\bf{Task-oriented example} \\  
    \midrule
    User input & Tokenized utterance  &does money buy happiness ? &i am looking for a cheap hotel . \\
    \midrule
    Belief state & [domain] slot \textit{value} & [chit] money happiness &[hotel] price cheap\\
    \midrule
    DB result &  \makecell[l]{A token indicated the number\\\ of candidate entities} &[db\_nore] &[db\_2]\\
    \midrule
    Act & [domain] [act] \textit{slot} &[chit] [chit\_act] & [hotel] [request] area\\
    \midrule
    Response & Tokenized utterance &\makecell[l]{depends on how much money\\ you spend on it .} &\makecell[l]{do you have a specific area you\\ want to stay in ?} \\
    
    \bottomrule
\end{tabular}
\caption{Unified dialogue data schema (\emph{Italicized} tokens are optional) and examples.}
\label{tab:data}
\end{table*}

We formulate UniDS as an auto-regressive (AR) language model and the dialogue response task is modeled as a sequence generation task.
A dialogue session at turn $t$ has the following components: user input $U_t$, belief state $B_t$, database search result $D_t$, system act $A_t$, and response $R_t$. Each component consists of tokens from a fixed vocabulary.
For turn $t$, the dialogue context $C_t$ is the concatenation of all the components of the previous dialogues as well as the user input at turn $t$: $C_t=[U_0, B_0, D_0, A_0, R_0, \cdots, R_{t-1}, U_{t}]$. Given the dialogue context $C_t$, UniDS first generates the belief state $B_t$:
\begin{equation}
B_t=\operatorname{UniDS}(C_t) \, ,
\end{equation}
and use it to search the database to get the search result $D_t$.
Then, UniDS generates the system act $A_t$ conditioned on the updated context by extending $C_t$ with $B_t$ and $D_t$:
\begin{equation}
A_t=\operatorname{UniDS}(C_t \oplus [B_t, D_t]) \, ,
\end{equation}
where $\oplus$ is the concatenation operator.
Lastly, the system response $R_t$ is generated, conditioned on the concatenation of all previous components:
\begin{equation}
R_t=\operatorname{UniDS}(C_t \oplus [B_t, D_t, A_t]) \, .
\end{equation}
The overview of the proposed unified dialogue system (UniDS) is illustrated in Figure~\ref{fig:overview}.

In the following sections, we will introduce the unified dialogue data schema and give the details of the training process of UniDS.

\subsection{Unified Dialogue Data Schema}

In the widely adopted end-to-end TOD pipeline, a dialogue session consists of a user input utterance, a belief state that represents the user intention, a database search result, a system act, and a system response \cite{YoungGTW13,YangLQ21}.
However, due to the diversity of chit-chat and the cost of manual annotation, most chit-chat dialogue systems do not assume the existence of the belief state, database result, nor system act \cite{abs-2006-16779,ZhangSGCBGGLD20}.
The inconsistency of data format between chit-chat and TOD hinders the implementation of a unified model.
To tackle this problem, we design a data schema with belief states, representation of database results, and system acts for chit-chat.
Table~\ref{tab:data} illustrates such unified data schema with examples. The following sections explain each component in detail.

\subsubsection{Belief state} Unified belief states are represented in the form of ``[domain] slot \textit{value}''. A belief state could have several domains, each containing several slot-value pairs.
For chit-chat, slots are the nouns extracted from the user utterance $U_t$; values are left empty.

\subsubsection{DB result} We use special tokens to represent the number of matched entities under the constraints of the belief state in the current turn.
For chit-chat, we use a token ``[db\_nore]'' to represent there are no matched entities.

\subsubsection{System act} System acts are represented as ``[domain] [act] \textit{slot}'' for TOD. ``[domain]'' is the same as in belief states. ``[act]'' denotes the type of action the system needs to perform. Following the ``domain-act'' pair, slots are optional. for chit-chat, we use the action ``[chit\_chat]'' to denote that the system needs to chit-chat with the user.

\subsection{Training Setup of UniDS}


The data for training UniDS is the set of dialogues mixed with chit-chat and TOD data which are pre-processed by the proposed unified data schema introduced before.
In this way, a processed dialog data sequence at turn $t$ for either TOD or chit-chat can be both represented as:
\begin{equation}
  X_t = [C_t,B_t,D_t,A_t,R_t] \, ,
\end{equation}
where $C_t$ is the dialog context, $B_t$ is the believe state, $D_t$ is the DB results, $A_t$ is the system action, $A_t$ is the system response, and the length of $X_t$ is $N$.

The training objective for UniDS is to maximize the joint probability of all tokens in $X_t$ computed in an auto-regressive manner as:
\begin{equation}\label{normal_weight}
\mathcal{L}=\sum^{N}_{i=1}-\log P(\boldsymbol{x}_{i}|\boldsymbol{x}_{<i}) \, ,
\end{equation}
where $\boldsymbol{x}_i$ is a token of $X_t$, and $\boldsymbol{x}_{<i}$ are the preceding tokens.

\subsubsection{Weighted cross entropy loss}
Chit-chat and TOD have different characteristics.
Chit-chat dialogues need to attract users to talk more, while TOD needs to complete the task as soon as possible.
Therefore, a model trained with the mixed dialogue data and pre-trained from chitchat might talk a large number of turns instead of efficiently completing the task.
Since ``entity recommendation'' acts are important for dialogue system to complete tasks efficiently, we propose a weighted cross-entropy loss ($L_{\boldsymbol{w}}$) as the training objective of UniDS. It assigns larger weights to tokens about the system's entity recommendation actions:
\begin{equation}\label{weight}
\mathcal{L}_{\boldsymbol{w}}=\sum^{N}_{i=1}-\boldsymbol{w}_i\log P(\boldsymbol{x}_{i}|\boldsymbol{x}_{<i}) \,
\end{equation}
If $\boldsymbol{x}_i$ is a token representing an entity recommendation act, the scalar weight $\boldsymbol{w}_i$ is set larger than $1$; for other regular tokens, $\boldsymbol{w}_i$ is set to $1$ by default.


\begin{table*}[!t]
\centering
\begin{tabular}{lccccc|cccc}
    \toprule
    \multirow{2}{*}{Model} & \# of para. &\multicolumn{4}{c}{Task-oriented Dialogue} &\multicolumn{4}{c}{Chit-chat}\\
     & \multicolumn{1}{c}{} & Inform & Success & BLEU & Combined & BLEU & Dist-1 & Dist-2 & AvgLen\\  
    \midrule
    UBAR-repro &82M &88.70 &78.40 &16.60 &100.15 &- &- &- &-\\
    \midrule
    UBAR-DialoGPT-12L &117M & \bf 89.40  & 75.10 & 16.93 & 99.18 &- &- &- &- \\
    DialoGPT-12L &117M &- &-  &-  &-  & 0.27 &\bf 6 &\bf 32 & 14.00 \\
    UniDS-12L &117M & 87.10 & \bf 77.00 & \bf 18.01 & \bf 100.06 & \bf 0.35 & \bf 6 & 30 &12.00\\
    \midrule
    UBAR-DialoGPT-24L &345M & 89.40 & 75.50 & 16.86 & 99.31  &- &- &- &-\\
    DialoGPT-24L &345M &- &-  &-  &- & 0.43  & \bf 7  & \bf 36  & 12.28 \\
    UniDS-24L &345M & \bf 90.30  & \bf 80.50 & \bf 18.72 & \bf 104.12 & \bf 0.45 & 6 & 35 & 14.62 \\
    \bottomrule
\end{tabular}
\caption{Automatic evaluations of UniDS with two model sizes over two types of dialogue datasets. All results are reported in percentage, except AvgLen. Best results are in \textbf{bold}. Only UniDS performs on both tasks.}
\label{overall}
\end{table*}

\section{Experiment}

\subsection{Datasets}
For training and evaluation, We mix chit-chat dialogues, exacted from Reddit,\footnote{https://www.reddit.com/} and MultiWOZ dataset \cite{BudzianowskiWTC18} with the proposed data schema.

\subsubsection{Chit-chat Dataset}
We derived chit-chat dialogue from Reddit dump.\footnote{\url{https://files.pushshift.io/reddit/comments/}}
The chit-chat training set and test set are extracted from the Reddit posts in 2017 and 2018 respectively, to ensure no overlapping.
To ensure the generation quality, we conduct a careful data cleaning.
A conversation will be filtered when (1) there is a URL in the utterance; (2) there is an utterance longer than $200$ words or less than $2$ words; (3) the dialogue contains ``[removed]" or ``[deleted]" tokens; (4) the number of utterances in the dialogue is less than 4.
Finally, we sample $8,438$ dialogues for training which is the same size as the training set of MultiWOZ. The validation set and test set contain $6,000$ dialogues and $8,320$ dialogues, respectively.

\subsubsection{MultiWOZ} For task-oriented dialogues, we adopt the publicly multi-domain goal-oriented MultiWOZ \cite{BudzianowskiWTC18}, which consists of $10,438$ dialogues spinning over seven domains (\textit{taxi, attraction, police, restaurant, train, hotel, hospital}).\footnote{We use MultiWOZ 2.0.}
The train/validation/test sets of MultiWOZ have $8438/1000/1000$ dialogues, respectively.
Each dialogue contains $1$ to $3$ domains.

\subsection{Baselines}
For chit-chat dialogue, we compare UniDS with \textbf{DialoGPT} \cite{ZhangSGCBGGLD20}.
For fair comparisons, we further fine-tune a 12-layer DialoGPT and a 24-layer DialoGPT with our chit-chat dialogue training set, which we refer to as DialoGPT-12L and DialoGPT-24L, respectively.

For TOD, we consider the state-of-the-art \textbf{UBAR} \cite{YangLQ21} model for end-to-end TOD system on top of DistilGPT2 \cite{abs-1910-01108}.
We report the results based on our reproduction of UBAR, referred to as UBAR-repro.
For a fair comparison with UniDS, we also fine-tune UBAR from 12 layers DialoGPT and 24 layers DialoGPT with MultiWOZ dataset, the fine-tuned models are denoted as UBAR-DialoGPT-12L and UBAR-DialoGPT-24L, respectively.

\subsection{Implementation Details}
UniDS and other baselines are implemented based on HuggingFace's Transformers \cite{abs-1910-03771}.
The max sequence length is $1024$ and sequences longer than 1024 are truncated from the head.
We use the AdamW optimizer \cite{Loshchilov2019DecoupledWD} and greedy decoding method for inference.
We report the main results of UniDS by setting the weight of entity recommendation tokens to $2$ (in Equation~\ref{weight})\footnote{The appendix gives discussions for other values of weight, but does not affect the overall conclusion.}. All models are trained on a single Tesla V100, and we perform a hyper-parameter search on batch size and learning rate.
The best model and hyper-parameter are selected through the performance on the validation set of MultiWOZ only.
\subsection{Evaluation Metrics}
For chit-chat dialogues, the BLEU score \cite{PapineniRWZ02} and the average length of the generated responses are reported.
Because of the diversity of chit-chat, BLEU may be difficult to reflect the quality of chit-chat responses, we also report distinct-1 and distinct-2 \cite{LiGBGD16} of generated dialogues, which is defined as the rate of distinct uni- and bi-grams in the generated sentences.
We also conduct a human evaluation on 50 randomly sampled test dialogues for two 24 layers models.
Three judges evaluate them in terms of relevance, informativeness, and how human-like the response is with a 3-point Likert-like scale \cite{Joshi2015LikertSE}. 

For TOD, we follow UBAR to use the following automatic metrics: \textbf{Inform} refers to the rate of the entities provided by a model are correct; \textbf{Success} measures the rate of a model has answered all the requested information; and \textbf{BLEU} to measure the fluency of generated responses.
Following \citet{Hosseini-AslMWY20, YangLQ21}, a combined score is computed as $(\text{Inform}+\text{Success})\times0.5+\text{BLEU}$ to measure the overall response quality.

\begin{table*}[!t]
\centering
\begin{tabular}{lcccc|cccc}
    \toprule
    \multirow{2}{*}{Model} &\multicolumn{4}{c}{Task-oriented Dialogue} &\multicolumn{4}{c}{Chit-chat}\\
     & Inform & Success & BLEU & Combined & BLEU & Dist-1 & Dist-2 & AvgLen\\  
    \midrule
    UniDS-12L &  87.10 &  77.00 & 18.01 &  100.06 & 0.35 &  6 &30 &12.00\\
    \ \ \ \ w/o chit-chat BS & 83.90 & 72.80 & 18.15 &96.50 &  0.37 & 5 & 29 &14.67\\
    \ \ \ \ w/o weighted loss &81.70 &71.20 &17.93 &94.38 &0.33 &  6 & 32 &14.29\\
    \midrule
    UniDS-24L &  90.30  &  80.50 & 18.72 &  104.12 & 0.45 &  6 &  35 & 14.62\\
    \ \ \ \ w/o chit-chat BS & 86.90 &78.50 &18.71 &101.41 &  0.49 &  6 &33 &15.29\\
    \ \ \ \ w/o weighted loss &85.60 &76.50 &  18.96 &100.01 &0.44 & 6 &34 &14.85\\
    \bottomrule
\end{tabular}
\caption{Ablation studies of automatic evaluations for UniDS.}
\label{ablation}
\end{table*}

\begin{figure*}[!t]
\centering
\includegraphics[width=1\textwidth]{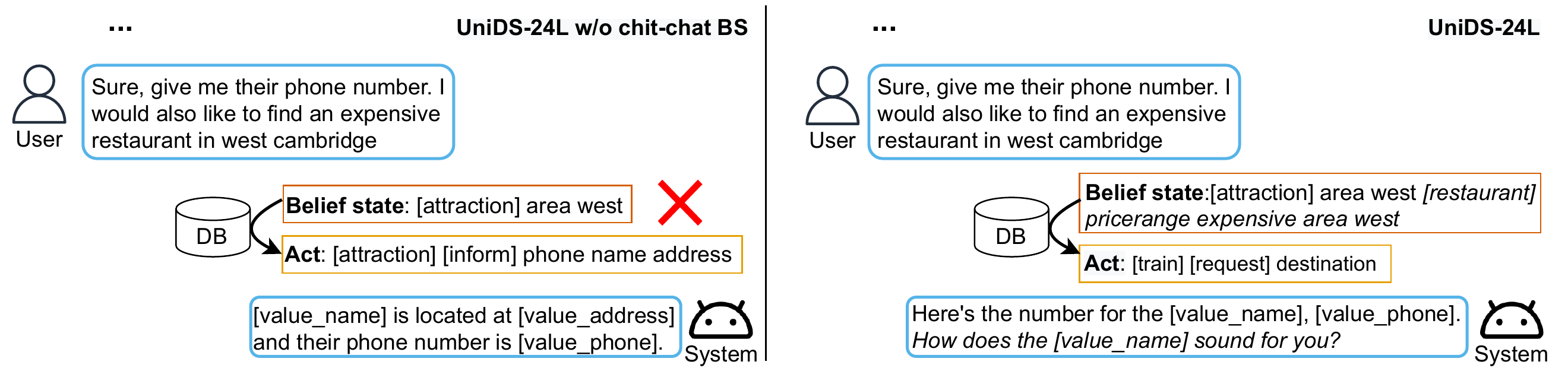} 
\caption{Task-oriented dialogue examples from UniDS w/o chit-chat BS and UniDS. UniDS w/o chit-chat BS does not extract the user intent of searching restaurants, but UniDS extracts this intent successfully (highlighted in italics).}
\label{noun}
\end{figure*}

\begin{table}[t]
\centering
\setlength{\tabcolsep}{4pt}
\begin{tabular}{cccc}
\toprule
 & DialoGPT-24L & Neutral & UniDS-24L \\ 
 &(Win \%)& (\% )&(Win \%)\\
\midrule
Relevance & 25.33 & 42.67 & \bfseries 32.00 \\
Informativeness & 29.33 &33.33 &\bfseries 37.34  \\
Human-like & 26.67 & 43.33 & \bfseries 30.00 \\
\bottomrule
\end{tabular}
\caption{Win rate [\%] between the UniDS-24L and DialoGPT-24L using three human evaluation metrics on chit-chat dialogues. ``Neutral'' means the generated responses of DialoGPT-24L and UniDS-24L are considered to have equal quality, and \textbf{bold} results represent the better response generated by the model.}
\label{tab:human_evaluation}
\end{table}

\subsection{Overall results}

Table~\ref{overall} presents the overall comparison results of automatic evaluation.
The first block shows the reproduced results of UBAR for chit-chat. The following two blocks are various baselines trained on 12 or 24 layers DialoGPT respectively. 
From these results, we have the following observations. 
\begin{enumerate}[i)]
\item For the chit-chat task, UniDS achieves comparable performance with DialoGPT. For the BLEU score, UniDS outperforms DialoGPT with 12L and 24L. On other metrics, UniDS is comparable with DialoGPT. This demonstrates that UniDS can still keep strong chit-chat ability even after training with the mixed dialogue data.
\item For the TOD task, UniDS achieves better performance than UBAR for the same parameter size. For both 12L and 24L DialoGPT, UniDS improves the BLEU score and the Combined score compared with UBAR. We believe this is because combining chit-chat dialogues for training helps the model to generate more fluent responses.
\end{enumerate}
Furthermore, we also provide the human evaluation results in Table~\ref{tab:human_evaluation}. UniDS is compared to DialoGPT regarding three dimensions for chit-chat dialogues. We could see that UniDS consistently wins the majority cases for all three aspects, including relevance, informativeness, and human-like.

\subsection{Analysis}

\subsubsection{Ablation Study}
\label{sec:ab}
In this experiment (c.f. Table~\ref{ablation}), we compare two simplified versions of UniDS to understand the effects of different components.
For comparison, we report the performance of 1) removing slots in belief state of chit-chat, denoted as ``UniDS w/o chit-chat BS'', and 2) replacing the weighted cross-entropy loss with a standard cross-entropy loss, denoted as ``UniDS w/o weighted loss''. Next, we elaborate our observations w.r.t. these two components.

\textbf{w/o chit-chat BS:} When removing the belief state of chit-chat dialogues, the performances of both UniDS-12L and UniDS-24L drop w.r.t. inform, success, and combined score for TOD.
We believe the reason is that the process of extracting the belief state can copy some keywords from the user utterance, and even extracting nouns as belief state for chit-chat is helpful for UniDS to learn this copy mechanism in the TOD task.
Taking the case in Figure~\ref{noun} as an example, UniDS w/o chit-chat BS (left) fails to extract the user's interest in searching restaurants, while UniDS (right) extracts the restaurant slot successfully. As a result, UniDS could recommend the right entities.
Furthermore, removing chit-chat BS does not degrade the performance of chit-chat.

\textbf{w/o weighted loss:} When replacing the weighted cross-entropy loss (c.f. Equation~\ref{weight}) in UniDS with standard cross-entropy loss, we observe a notable drop w.r.t. inform, success, and combined score in task-oriented metrics.
These results demonstrate that giving more attention to entity recommendation acts helps the task completion capability.
Moreover, dropping the weight loss does not affect the performance of chit-chat much.

Overall, we contend both ``chit-chat BS'' and ``weighted loss'' are beneficial for task-oriented dialogues without degrading the chit-chat capability.

\subsubsection{Analysis of Switching Ability}

In real-world scenarios, it is common and natural for users to switch between chit-chat and task-oriented dialogues.
Therefore, we demonstrate that the proposed UniDS has the ability to switch between two dialogue tasks.
To simulate the scenario of dialogue switching, we consider two setups: (1) having two turns of chit-chat dialogues before the start of a task-oriented dialogue and (2) pre-pending two turns of task-oriented dialogues at the beginning of a chit-chat dialogue.
To evaluate the model's ability to switch between two types of dialogues, we propose a metric, called \textbf{Switch-$n$}, which is defined as the success rate of model switching response type within the first $n$ turns after the switching.
Additionally, we also report the model performance \textbf{after} the switching.

\begin{table}[!t]
\small
\setlength{\tabcolsep}{3pt}
\centering
        \begin{tabular}{ccccccc}
            \toprule
            UniDS & Inf. & Succ. & BLEU & Comb. & Switch-1 & Switch-2\\
            \midrule
            12L &84.60 &72.00 &11.72 &90.02 & 65.8 & 99.5 (+33.7)\\
            24L &85.30 &75.70 &12.44 &92.94 & 64.4 & 99.2 (+34.8)\\
            \bottomrule
        \end{tabular}
        \caption{Switching performance of UniDS when having 2 turns chit-chat dialogues before task-orientated dialogues. Numbers in brackets indicates the exactly switching rate at the 2nd turn.}
       \label{tab:switch1}
\end{table}

\begin{table}[!t]
\small
\setlength{\tabcolsep}{3pt}
\centering

        \begin{tabular}{@{}ccccccc@{}}
            \toprule
            UniDS & BLEU & Dist-1 & Dist-2 & AvgLen & Switch-1 & Switch-2\\ 
            \midrule
            12L  &0.22 & 4 &19 & 14.15 & 31.8 & 98.9 (+67.1) \\
            24L  &0.34 & 6 &31 & 16.18 & 37.0 & 96.6 (+59.6) \\
            \bottomrule
        \end{tabular}
        \caption{Switching performance of UniDS when pre-pending 2 turns task-oriented dialogues before chit-chat.}
       \label{tab:switch2}
\end{table}

Table~\ref{tab:switch1} and Table~\ref{tab:switch2} present the results of the two switching setups, and we have the following observations:
\begin{enumerate}[(i)]
\item
It is not surprising that adding switching tasks for both chit-chat and TOD degrades the performance of UniDS, as the added 2 turns of switching utterances introduce irrelevant content, which distracts the model.
However, focusing on the switching task, we observe that for almost 98\% of cases, UniDS can success in dialogue task switching, from chit-chat to TOD and vice versa, within the first two turns (Switch-$1$ and Switch-$2$).
This demonstrates UniDS has a good ability to switch between two types of dialogue tasks.
\item When switching from task-oriented dialogues to chit-chat dialogues, the value of Switch-$1$ is relatively low, this may because our model tends to confirm user intents or give a transitional response rather than switch to chit-chat mode immediately. As the case shown in Table~\ref{switch1}, when the user switches from TOD to chit-chat, UniDS gives a chatty response and thanks the user for using its services.
\end{enumerate}

\begin{table}[!t]
\centering
\begin{tabularx}{1\linewidth}{X}
\toprule
\textbf{User@Turn $t$:} \\
\textcolor{cyan}{Will the Red Tesla get me to the restaurant for my 16:15 reservation?}\\
\textbf{Response generated by UniDS:}\\
\textcolor{cyan}{Yes, your taxi will pick you up at all saints church and arrive at tang chinese by 16:15. Is there anything else I can do for you today?}\\
\hdashline
\textbf{User@Turn $t+1$:}\\
\textcolor{cyan}{Ok sounds great. That is all I need. Thank you!} \\
\textbf{Response generated by UniDS:}\\
\textcolor{cyan}{Thank you for using our services.}\\
\hdashline
\textbf{User@Turn $t+2$:} \\
\textcolor{teal}{\textit{How's your new year's resolution holding up?}}\\
\textbf{Response generated by UniDS:}\\
\textcolor{teal}{\textit{It's holding up just fine. Thanks for inquiring with us.}}\\
\bottomrule
\end{tabularx}
\caption{Example of UniDS when switching from the \textcolor{cyan}{task-oriented dialogue} to \textcolor{teal}{\textit{chit-chat}}. UniDS gives a chatty response and thanks the user for using its services. Dialogue history is omitted.}
\label{switch1}
\end{table}

\begin{table}[t]
\setlength{\tabcolsep}{3.5pt}
\centering
    \small
    \begin{tabular}{lccc}
    \toprule
        Model &  Base & 1 turn & 2 turns \\
    \midrule
        UBAR-DialoGPT-12L & 99.18  & 93.76 (-5.42) & 88.14 (-11.04) \\
        UniDS-12L         & 100.06 & 96.13 (-3.93) & 91.42 (-8.64) \\
    \midrule
        UBAR-DialoGPT-24L & 99.31  & 93.08 (-6.23) & 88.67 (-10.64) \\
        UniDS-24L         & 104.12 & 100.71 (-3.41) & 95.68 (-8.44)\\
    \bottomrule
    \end{tabular}

    \caption{Combined score over TOD dataset for robustness test by inserting 1 and 2 turns of task-irrelavant utterances. Full results are presented in Appendix.}
    \label{rob}
\end{table}

\subsubsection{Robustness Study}

\begin{figure*}[!t]
\centering
\includegraphics[width=\textwidth]{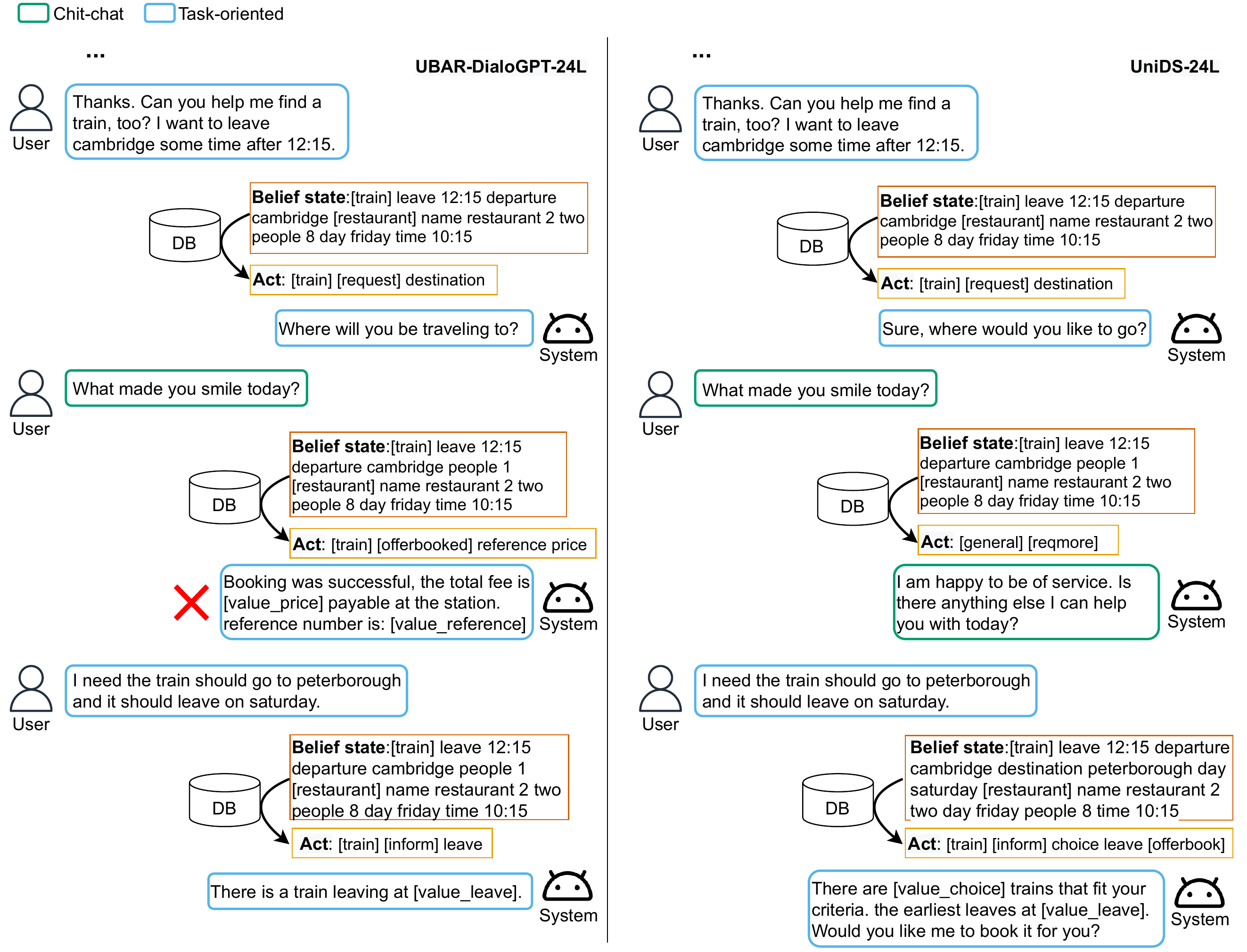}
\caption{Examples of UBAR-DialoGPT-24L and UniDS-24L when inserting a task-irrelevant utterance in a task-oriented dialogue. UBAR-DialoGPT reserves a train for the user randomly, which makes the task failed because the user intent is incomplete; while UniDS keeps the previous belief state and gives a chatty response. When the user returns to the TOD, UniDS could continue with the task.} 
\label{rob_case}
\end{figure*}

Many real-world dialogue systems need real-time speech recognition to interact with users, which is easily interfered by background noise from the background environment (e.g. other people and devices).
Therefore, we analyze the robustness of UniDS and UBAR by inserting several turns of irrelevant utterances into the TOD, and we evaluate the model performance against such noise.

As observed in Table~\ref{rob}, both UniDS and UBAR-DialoGPT drops on the combined score when only one turn of chit-chat dialogue is inserted. However, UniDS drop less than UBAR-DialoGPT (4 vs. 6 points).
Similarly, when two turns of chit-chat are inserted into TOD, UniDS drops about 8.4 points, and UBAR-DialoGPT drops about 11 points on the combined score.
These results demonstrate that our UniDS has stronger robustness to such task-irrelevant noise than UBAR-DialoGPT.
We present an interesting case in Figure~\ref{rob_case}.
When giving a task-irrelevant utterance, UBAR-DialoGPT reserves a train for the user randomly, which makes the task failed because the user intent is incomplete, while UniDS keeps the previous belief state and gives a chatty response.
When the user returns to the TOD, UniDS can continue with the task.

\section{Conclusion}
This paper proposes a unified dialogue system (UniDS) that can handle both end-to-end chit-chat and task-oriented dialogue in a joint framework.
We present a unified dialogue data schema that contains belief state, database result, and system act for two types of dialogues.
In this way, we could utilize the mixed dialogue data to further train the chit-chat dialogue system, and enable the trained model to handle both chit-chat and task-oriented dialogues.
To our best knowledge, this is the first study towards an end-to-end unified dialogue system.

Experiments show that UniDS performs comparably with state-of-the-art chit-chat dialogue systems and outperforms the state-of-the-art task-oriented dialogue systems without adding extra parameters.
More importantly, the proposed UniDS achieves better switching ability and robustness than previous task-oriented dialogue systems.
As an initial attempt, our explorations may inspire future studies towards building more capable and more intelligent dialog systems.

\bibliography{ref}

\begin{thebibliography}{23}
\providecommand{\natexlab}[1]{#1}
\providecommand{\url}[1]{\texttt{#1}}
\providecommand{\urlprefix}{URL }
\expandafter\ifx\csname urlstyle\endcsname\relax
  \providecommand{\doi}[1]{doi:\discretionary{}{}{}#1}\else
  \providecommand{\doi}{doi:\discretionary{}{}{}\begingroup
  \urlstyle{rm}\Url}\fi

\bibitem[{Adiwardana et~al.(2020)Adiwardana, Luong, So, Hall, Fiedel,
  Thoppilan, Yang, Kulshreshtha, Nemade, Lu, and Le}]{abs-2001-09977}
Adiwardana, D.; Luong, M.; So, D.~R.; Hall, J.; Fiedel, N.; Thoppilan, R.;
  Yang, Z.; Kulshreshtha, A.; Nemade, G.; Lu, Y.; and Le, Q.~V. 2020.
\newblock Towards a Human-like Open-Domain Chatbot.
\newblock \emph{CoRR} abs/2001.09977.

\bibitem[{Bao et~al.(2020)Bao, He, Wang, Wu, Wang, Wu, Guo, Liu, and
  Xu}]{abs-2006-16779}
Bao, S.; He, H.; Wang, F.; Wu, H.; Wang, H.; Wu, W.; Guo, Z.; Liu, Z.; and Xu,
  X. 2020.
\newblock {PLATO-2:} Towards Building an Open-Domain Chatbot via Curriculum
  Learning.
\newblock \emph{CoRR} abs/2006.16779.

\bibitem[{Budzianowski et~al.(2018)Budzianowski, Wen, Tseng, Casanueva, Ultes,
  Ramadan, and Gasic}]{BudzianowskiWTC18}
Budzianowski, P.; Wen, T.-H.; Tseng, B.-H.; Casanueva, I.; Ultes, S.; Ramadan,
  O.; and Gasic, M. 2018.
\newblock MultiWOZ-A Large-Scale Multi-Domain Wizard-of-Oz Dataset for
  Task-Oriented Dialogue Modelling.
\newblock In \emph{Proceedings of the 2018 Conference on Empirical Methods in
  Natural Language Processing}, 5016--5026.

\bibitem[{Hosseini{-}Asl et~al.(2020)Hosseini{-}Asl, McCann, Wu, Yavuz, and
  Socher}]{Hosseini-AslMWY20}
Hosseini{-}Asl, E.; McCann, B.; Wu, C.; Yavuz, S.; and Socher, R. 2020.
\newblock A Simple Language Model for Task-Oriented Dialogue.
\newblock In \emph{Advances in Neural Information Processing Systems 33: Annual
  Conference on Neural Information Processing Systems 2020, NeurIPS 2020,
  December 6-12, 2020, virtual}.

\bibitem[{Joshi et~al.(2015)Joshi, Kale, Chandel, and Pal}]{Joshi2015LikertSE}
Joshi, A.; Kale, S.; Chandel, S.; and Pal, D. 2015.
\newblock Likert Scale: Explored and Explained.
\newblock \emph{British Journal of Applied Science and Technology} 7: 396--403.

\bibitem[{Kaiser et~al.(2017)Kaiser, Gomez, Shazeer, Vaswani, Parmar, Jones,
  and Uszkoreit}]{KaiserGSVPJU17}
Kaiser, L.; Gomez, A.~N.; Shazeer, N.; Vaswani, A.; Parmar, N.; Jones, L.; and
  Uszkoreit, J. 2017.
\newblock One Model To Learn Them All.
\newblock \emph{CoRR} abs/1706.05137.

\bibitem[{Li et~al.(2016)Li, Galley, Brockett, Gao, and Dolan}]{LiGBGD16}
Li, J.; Galley, M.; Brockett, C.; Gao, J.; and Dolan, B. 2016.
\newblock A Diversity-Promoting Objective Function for Neural Conversation
  Models.
\newblock In Knight, K.; Nenkova, A.; and Rambow, O., eds., \emph{{NAACL} {HLT}
  2016, The 2016 Conference of the North American Chapter of the Association
  for Computational Linguistics: Human Language Technologies, San Diego
  California, USA, June 12-17, 2016}, 110--119. The Association for
  Computational Linguistics.

\bibitem[{Lin et~al.(2021)Lin, Madotto, Bang, and Fung}]{LinMBF21}
Lin, Z.; Madotto, A.; Bang, Y.; and Fung, P. 2021.
\newblock The Adapter-Bot: All-In-One Controllable Conversational Model.
\newblock In \emph{Thirty-Fifth {AAAI} Conference on Artificial Intelligence,
  {AAAI} 2021, Thirty-Third Conference on Innovative Applications of Artificial
  Intelligence, {IAAI} 2021, The Eleventh Symposium on Educational Advances in
  Artificial Intelligence, {EAAI} 2021, Virtual Event, February 2-9, 2021},
  16081--16083. {AAAI} Press.

\bibitem[{Loshchilov and Hutter(2019)}]{Loshchilov2019DecoupledWD}
Loshchilov, I.; and Hutter, F. 2019.
\newblock Decoupled Weight Decay Regularization.
\newblock In \emph{ICLR}.

\bibitem[{Madotto et~al.(2020)Madotto, Lin, Wu, Shin, and
  Fung}]{abs-2001-01871}
Madotto, A.; Lin, Z.; Wu, C.; Shin, J.; and Fung, P. 2020.
\newblock Attention over Parameters for Dialogue Systems.
\newblock \emph{CoRR} abs/2001.01871.

\bibitem[{Papineni et~al.(2002)Papineni, Roukos, Ward, and Zhu}]{PapineniRWZ02}
Papineni, K.; Roukos, S.; Ward, T.; and Zhu, W. 2002.
\newblock Bleu: a Method for Automatic Evaluation of Machine Translation.
\newblock In \emph{Proceedings of the 40th Annual Meeting of the Association
  for Computational Linguistics, July 6-12, 2002, Philadelphia, PA, {USA}},
  311--318. {ACL}.

\bibitem[{Peng et~al.(2020)Peng, Li, Li, Shayandeh, Liden, and
  Gao}]{abs-2005-05298}
Peng, B.; Li, C.; Li, J.; Shayandeh, S.; Liden, L.; and Gao, J. 2020.
\newblock {SOLOIST:} Few-shot Task-Oriented Dialog with {A} Single Pre-trained
  Auto-regressive Model.
\newblock \emph{CoRR} abs/2005.05298.

\bibitem[{Radford et~al.(2019)Radford, Wu, Child, Luan, Amodei, Sutskever
  et~al.}]{radford2019language}
Radford, A.; Wu, J.; Child, R.; Luan, D.; Amodei, D.; Sutskever, I.; et~al.
  2019.
\newblock Language models are unsupervised multitask learners.
\newblock \emph{OpenAI blog} 1(8): 9.

\bibitem[{Roller et~al.(2021)Roller, Dinan, Goyal, Ju, Williamson, Liu, Xu,
  Ott, Smith, Boureau, and Weston}]{RollerDGJWLXOSB21}
Roller, S.; Dinan, E.; Goyal, N.; Ju, D.; Williamson, M.; Liu, Y.; Xu, J.; Ott,
  M.; Smith, E.~M.; Boureau, Y.; and Weston, J. 2021.
\newblock Recipes for Building an Open-Domain Chatbot.
\newblock In \emph{Proceedings of the 16th Conference of the European Chapter
  of the Association for Computational Linguistics: Main Volume, {EACL} 2021,
  Online, April 19 - 23, 2021}, 300--325. Association for Computational
  Linguistics.

\bibitem[{Sanh et~al.(2019)Sanh, Debut, Chaumond, and Wolf}]{abs-1910-01108}
Sanh, V.; Debut, L.; Chaumond, J.; and Wolf, T. 2019.
\newblock DistilBERT, a distilled version of {BERT:} smaller, faster, cheaper
  and lighter.
\newblock \emph{CoRR} abs/1910.01108.

\bibitem[{Sun et~al.(2021)Sun, Moon, Crook, Roller, Silvert, Liu, Wang, Liu,
  Cho, and Cardie}]{SunMCRSLWLCC21}
Sun, K.; Moon, S.; Crook, P.~A.; Roller, S.; Silvert, B.; Liu, B.; Wang, Z.;
  Liu, H.; Cho, E.; and Cardie, C. 2021.
\newblock Adding Chit-Chat to Enhance Task-Oriented Dialogues.
\newblock In \emph{Proceedings of the 2021 Conference of the North American
  Chapter of the Association for Computational Linguistics: Human Language
  Technologies, {NAACL-HLT} 2021, Online, June 6-11, 2021}, 1570--1583.
  Association for Computational Linguistics.

\bibitem[{Wolf et~al.(2019)Wolf, Debut, Sanh, Chaumond, Delangue, Moi, Cistac,
  Rault, Louf, Funtowicz, and Brew}]{abs-1910-03771}
Wolf, T.; Debut, L.; Sanh, V.; Chaumond, J.; Delangue, C.; Moi, A.; Cistac, P.;
  Rault, T.; Louf, R.; Funtowicz, M.; and Brew, J. 2019.
\newblock HuggingFace's Transformers: State-of-the-art Natural Language
  Processing.
\newblock \emph{CoRR} abs/1910.03771.

\bibitem[{Yang, Li, and Quan(2021)}]{YangLQ21}
Yang, Y.; Li, Y.; and Quan, X. 2021.
\newblock {UBAR:} Towards Fully End-to-End Task-Oriented Dialog System with
  {GPT-2}.
\newblock In \emph{Thirty-Fifth {AAAI} Conference on Artificial Intelligence,
  {AAAI} 2021, Thirty-Third Conference on Innovative Applications of Artificial
  Intelligence, {IAAI} 2021, The Eleventh Symposium on Educational Advances in
  Artificial Intelligence, {EAAI} 2021, Virtual Event, February 2-9, 2021},
  14230--14238. {AAAI} Press.

\bibitem[{Young et~al.(2013)Young, Gasic, Thomson, and Williams}]{YoungGTW13}
Young, S.~J.; Gasic, M.; Thomson, B.; and Williams, J.~D. 2013.
\newblock POMDP-Based Statistical Spoken Dialog Systems: {A} Review.
\newblock \emph{Proc. {IEEE}} 101(5): 1160--1179.

\bibitem[{Zhang et~al.(2020)Zhang, Sun, Galley, Chen, Brockett, Gao, Gao, Liu,
  and Dolan}]{ZhangSGCBGGLD20}
Zhang, Y.; Sun, S.; Galley, M.; Chen, Y.; Brockett, C.; Gao, X.; Gao, J.; Liu,
  J.; and Dolan, B. 2020.
\newblock {DIALOGPT} : Large-Scale Generative Pre-training for Conversational
  Response Generation.
\newblock In \emph{Proceedings of the 58th Annual Meeting of the Association
  for Computational Linguistics: System Demonstrations, {ACL} 2020, Online,
  July 5-10, 2020}, 270--278. Association for Computational Linguistics.

\bibitem[{Zhao et~al.(2017)Zhao, Lu, Lee, and Eskenazi}]{ZhaoLLE17}
Zhao, T.; Lu, A.; Lee, K.; and Eskenazi, M. 2017.
\newblock Generative Encoder-Decoder Models for Task-Oriented Spoken Dialog
  Systems with Chatting Capability.
\newblock In \emph{Proceedings of the 18th Annual SIGdial Meeting on Discourse
  and Dialogue, Saarbr{\"{u}}cken, Germany, August 15-17, 2017}, 27--36.
  Association for Computational Linguistics.

\bibitem[{Zhao et~al.(2020)Zhao, Xiao, Zhong, Yao, and Chen}]{ZhaoXZYC20}
Zhao, X.; Xiao, F.; Zhong, H.; Yao, J.; and Chen, H. 2020.
\newblock Condition Aware and Revise Transformer for Question Answering.
\newblock In Huang, Y.; King, I.; Liu, T.; and van Steen, M., eds., \emph{{WWW}
  '20: The Web Conference 2020, Taipei, Taiwan, April 20-24, 2020}, 2377--2387.
  {ACM} / {IW3C2}.

\bibitem[{Zhou et~al.(2018)Zhou, Huang, Zhang, Zhu, and Liu}]{ZhouHZZL18}
Zhou, H.; Huang, M.; Zhang, T.; Zhu, X.; and Liu, B. 2018.
\newblock Emotional Chatting Machine: Emotional Conversation Generation with
  Internal and External Memory.
\newblock In McIlraith, S.~A.; and Weinberger, K.~Q., eds., \emph{Proceedings
  of the Thirty-Second {AAAI} Conference on Artificial Intelligence, (AAAI-18),
  the 30th innovative Applications of Artificial Intelligence (IAAI-18), and
  the 8th {AAAI} Symposium on Educational Advances in Artificial Intelligence
  (EAAI-18), New Orleans, Louisiana, USA, February 2-7, 2018}, 730--739. {AAAI}
  Press.

\end{thebibliography}


\appendix

\begin{table}[h]
\centering
\section{Analysis of the value of $w$ in Loss Function}
\begin{tabular}{lcccc|cccc}
    \toprule
    \multirow{2}{*}{Model} &\multicolumn{4}{c}{Task-oriented} &\multicolumn{4}{c}{Chit-chat}\\
     & Inform & Success & BLEU & Combined & BLEU & Dist-1 & Dist-2 & AvgLen\\  
    \midrule
    UniDS-12L, $w=1$ &81.70 &71.20 &17.93 &94.38 &0.33 &\textbf{6} &\textbf{32} &14.29\\
    UniDS-12L, $w=2$ &87.10 &77.00 &\textbf{18.01} &100.06 &\textbf{0.35} &\textbf{6} &30 &12.00\\
    UniDS-12L, $w=3$ &90.50 &78.90 &17.65 &102.35 &0.32 &\textbf{6} &30 &13.47\\
    UniDS-12L, $w=4$ &91.30 &79.40 &17.90 &103.25 &\textbf{0.35} &\textbf{6} &30 &13.52\\
    UniDS-12L, $w=5$ &\textbf{92.40} &\textbf{80.60} &17.72 &\textbf{104.22} &0.34 &5 &30 &13.51\\
    \bottomrule
\end{tabular}
\caption{Performance comparison of UniDS-12L trained with different $w$}
\label{weight}
\end{table}
To analyse the influence of the value of $w$ on the performance of UniDS, we change the value of $w$ in Equation \ref{weight} to train UniDS-12L. As we can see from Table \ref{weight}, the performance of UniDS in task-oriented dialogues increased as $w$ increased. However, the performance of UniDS in chit-chat shows a decreasing trend.

\section{Full results on Robustness}
\label{sec:full_rob}

\begin{table}[!ht]
\centering
    \small
    
    
    \begin{subtable}[!htbp]{\linewidth}
        \begin{tabular}{l|cccc}
            \toprule
            Model & Inf. & Succ. & BLEU & Comb.\\
            \midrule
            UBAR-DialoGPT-12L & \bf 87.30 &72.60 &13.81 &93.76 \\
            UniDS-12L &86.60 & \bf 76.10 & \bf 14.78 & \bf 96.13 \\
            \midrule
            UBAR-DialoGPT-24L &86.60 &72.10 &13.73 &93.08\\
            UniDS-24L & \bf 90.20 & \bf 80.10 & \bf 15.56 & \bf 100.71 \\
            \bottomrule
        \end{tabular}
        \caption{Inserting {\bf 1 turn} chit-chat dialogue into random turn of task-orientated dialogues.}
      \label{tab:full_rob1}
    \end{subtable}
    \hfill
    \begin{subtable}[!htbp]{\linewidth}
        \begin{tabular}{l|cccc}
            \toprule
            Model & Inf. & Succ. & BLEU & Comb.\\ 
            \midrule
            UBAR-DialoGPT-12L &84.10 &69.50 &11.34 &88.14\\
            UniDS-12L & \bf 85.20 & \bf 73.50 & \bf 12.07 & \bf 91.42 \\
            \midrule
            UBAR-DialoGPT-24L &85.10 &70.00 &11.12 &88.67\\
            UniDS-24L & \bf 88.00 & \bf 77.40 & \bf 12.98 & \bf 95.68\\
            \bottomrule
        \end{tabular}
        \caption{Inserting {\bf 2 turns} chit-chat dialogues into random turn of task-orientated dialogues.}
      \label{tab:full_rob2}
    \end{subtable}
    \caption{Full results of robustness tests. The average turns of MultiWOZ is 6.84. The average turns of MultiWOZ is 6.84.}
    \label{full_rob}
\end{table}

\end{document}